\documentclass[
]{ceurart}

\usepackage{listings}
\usepackage{colortbl}
\usepackage{amsmath}
\begin{document}

\copyrightyear{2023}
\copyrightclause{Copyright for this paper by its authors. Use permitted under Creative Commons License Attribution 4.0 International (CC BY 4.0)}
%\copyrightclause{Copyright for this paper by its authors.
 % Use permitted under Creative Commons License Attribution 4.0
%  International (CC BY 4.0).}

\conference{Scholarly QALD at ISWC, November 6-10, 2023, Athens, Greece}

\title{PSYCHIC: A Neuro-Symbolic Framework for Knowledge Graph Question-Answering Grounding}

\tnotetext[1]{Scholarly QALD at ISWC, November 6-10, 2023, Athens, Greece}

\author[1,2]{Hanna Abi Akl}[%
orcid=0000-0001-9829-7401,
email=hanna.abi-akl@dsti.institute]
\address[1]{Data ScienceTech Institute (DSTI),
 4 Rue de la Collégiale 75005 Paris, France}
\address[2]{Université Côte d’Azur, Inria, CNRS, I3S}

\cortext[1]{Corresponding author.}

\begin{abstract}
  The Scholarly Question Answering over Linked Data (Scholarly QALD) \footnote{https://kgqa.github.io/scholarly-QALD-challenge/2023/} at The International Semantic Web Conference (ISWC) 2023 challenge presents two sub-tasks to tackle question answering (QA) over knowledge graphs (KGs). We answer the KGQA over DBLP (DBLP-QUAD) task by proposing a neuro-symbolic (NS) framework based on PSYCHIC \footnote{https://huggingface.co/HannaAbiAkl/psychic}, an extractive QA model capable of identifying the query and entities related to a KG question. Our system achieved a F1 score of 00.18\% on question answering and came in third place for entity linking (EL) with a score of 71.00\%.
\end{abstract}

\begin{keywords}
  Question Answering \sep
  Knowledge Graphs \sep
  Neuro-Symbolic Artificial Intelligence \sep
  Entity Linking \sep
  Linked Data \sep
  Language Models
\end{keywords}

\maketitle

\section{Introduction}
Knowledge veracity has always been one of the key topics of the Web \cite{peng2023knowledge}. The ability to present users with a body of factual information to refer to is a challenge that continues to defy the growth of the web \cite{peng2023knowledge}. In particular, the plethora of available information is plagued by the non-structured nature of this knowledge \cite{peng2023knowledge}.

One way the Semantic Web community addresses this issue is by constructing KGs which are structured repertoires of entities and relationships \cite{peng2023knowledge}. These graphs encapsulate factual knowledge and allow users to retrieve it by navigating their different connections \cite{peng2023knowledge}. The DBLP computer science bibliography is an example of such an effort that provides open bibliographic information on major computer science journals and proceedings \cite{banerjee2023dblp}.

Aside from consolidating bodies of information, KGs provide a platform for information retrieval \cite{peng2023knowledge}. In artificial intelligence (AI), question answering refers to the task of asking an AI agent a question and receiving an answer in return \cite{ishwari2019advances}. Large Language Models (LLMs) are popular agents that fill this role well due to their nature of ingesting huge amounts of information \cite{pan2023large}. However, this also makes them prone to giving out misinformation based on their inability to disseminate correct from incorrect knowledge \cite{pan2023large}.

Integrating LLMs and KGs has emerged as a solution to mitigate the shortcomings of large language models \cite{pan2023large}. By querying a factual base of information, LLMs gain a way to validate their responses before sending them back to users \cite{pan2023large}. It is in this scope that the Scholarly QALD challenge presents its two sub-tasks, DBLP-QUAD and SciQA. We focus on DBLP-QUAD, a QA task over the DBLP \footnote{https://blog.dblp.org/2022/03/02/dblp-in-rdf/} KG. We distinguish two parts in the challenge: question answering, which requires participants to propose systems capable of retrieving specific information from the KG to answer a question, and entity linking, which requires participants to retrieve the list of entities related to a question.

In this paper, we show how a NS system can handle the tasks of QA and EL over a KG. The rest of the paper is organized as follows. In section 2, we discuss some of the related work. In section 3, we present the experimental setup. In section 4, we discuss the results. Finally, we present our conclusions in section 5.

\section{Related Work \label{sec2}}
This section reviews some of the proposed systems designed for QA and EL over KGs.

\subsection{Question answering over knowledge graphs}

Several techniques have emerged to tackle the problem of QA over KGs. \citeauthor{zheng2019question} make use of structured query patterns which involve identifying query graph candidates in the KG using EL and disambiguation and transforming them to SPARQL queries \cite{zheng2019question} to answer questions. \citeauthor{pramanik2021uniqorn} employ a similar approach by proposing a model that draws from relevant RDF triples to generate all possible query context graphs from which queries are created to answer natural language questions. Their findings result in an advancement in graph-based methods but prove their approach to be highly noisy \cite{pramanik2021uniqorn}. \citeauthor{sima2022bio} improve on this approach by leveraging graph algorithms to identify and rank domain-specific query candidates based on the node centrality of the relevant entities. \citeauthor{zheng2018question} go a step further in this direction by integrating semantic parsers to their query templates to refine the query-generation process. By aligning both natural language questions and query templates, they prove they can effectively answer complex and detailed questions \cite{zheng2018question}.

\citeauthor{nikas2021open} combine graph-based techniques with neural-based methods to create a NS QA system. In their work, they train a DistilBERT model based on an expected answer type to handle different kinds of questions \cite{nikas2021open}. The neural network also leverages SPARQL queries to gather facts for entity enrichment and boost its performance in extracting the correct answer from the provided context \cite{nikas2021open}. \citeauthor{cm2022question} propose a similar pipeline by replacing the SPARQL endpoint with a supervised BERT model that performs relation extraction to add context information to the system.

\citeauthor{diomedi2107question} propose a different approach by using neural machine translation to map the questions in natural language to SPARQL templates directly. They show that this method, coupled with tree-based entity disambiguation techniques, turns the QA problem to a slot-filling task and outperforms vanilla deep learning (DL) models \cite{diomedi2107question}. \citeauthor{aghaei2022question} leverage a similar slot-filling pipeline on a domain-specific KG to demonstrate that it can reliably learn the pattern structures of the domain queries.

In their work, \citeauthor{mavromatis2022rearev} and \citeauthor{li2021improving} leverage graph networks to compute graph embeddings and compare them with the input question embeddings to target the correct answer entities. \citeauthor{dutt2022perkgqa} utilize graph convolution networks to score questions based on similarity and derive their corresponding graph paths to answer new questions.

\citeauthor{saxena2020improving} and \citeauthor{zuo2023improving} demonstrate how using KG embeddings can help solve multi-hop questions. \citeauthor{rony2022tree} propose a system whereby KG embeddings are stored in a vector database for faster retrieval and improved performance in answering complex questions.

\subsection{Entity linking over knowledge graphs}

\citeauthor{shi2023knowledge} present a survey on the techniques seen in EL over KGs. They regroup approaches into rule-based, machine learning (ML) and DL EL methods \cite{shi2023knowledge}. In their work, \citeauthor{dubey2018earl} use rule-based methods by leveraging global traveling salesman approximate solver algorithms to disambiguate entities in large KGs. \citeauthor{steinmetz2023entity} employ abstract meaning representation, a series of text dependency parsing rules, to identify and group sentences having the same meaning but different structures. Using this technique, they perform data augmentation to enrich entity representation and achieve better performance on the EL task \cite{steinmetz2023entity}. In a similar fashion, \citeauthor{radhakrishnan2018elden} use co-occurrences from large corpora to enrich existing KGs with entity information and create dense KGs as a basis for EL.

\citeauthor{thawanientity} employ ML techniques by combining TF-IDF with Wikidata entries to generate feature vectors and score entity candidates accordingly. \citeauthor{li2022improving} make use of translating embeddings to encode entity-entity relationships and combine them with entity-relation embeddings to get better performance over KGs.

From a DL perspective, \citeauthor{huang2020entity} show that using BERT to calculate sentence embeddings over entities outperforms rule-based entity enrichment approaches. \citeauthor{banerjee2020pnel} improve on this approach by concatenating entity embeddings with context using FastText embeddings in a pointer network entity linker structure designed to represent entities in a dense vector space and identify the correct one for any input entity.

Finally, \citeauthor{ding2021jel} propose a NS approach based on EL using SpaCy embeddings in a case-based reasoning by computing the cosine score between input entity embeddings and KG entity embeddings. The final entity is computed automatically according to the best matching embedding or manually labeled based on a threshold cosine score \cite{ding2021jel}. \citeauthor{diomedi2022entity} experiment with a pipeline that integrates rule-based entity matching over DBpedia and Wikidata KGs and neural network models to associate the resulting entities to a fixed set from a designed KG.

\section{Experiments \label{sec3}}
This section describes the framework for our experiments in terms of data, system and training process.

\subsection{Dataset description}

The dataset considered for this shared task is divided in two parts: the DBLP-QUAD \footnote{https://huggingface.co/datasets/awalesushil/DBLP-QuAD} dataset which consists of 10000 question-SPARQL pairs and is answerable over the DBLP KG, and a dataset of 500 questions retaining the same format provided by the task organizers which will be referred to as seed data. We provide details for both datasets in the following subsections.

\subsubsection{DBLP-QUAD data}

DBLP-QUAD is a scholarly KGQA dataset with 10,000 question-SPARQL query pairs targeting the DBLP KG. DBLP-QUAD was created using the OVERNIGHT approach where logical forms are first generated from a KG. Canonical questions are then generated from these logical forms. The dataset is split into 7,000 training, 1,000 validation and 2,000 test questions. Each question-SPARQL pair consists of the following data fields: the id of the question (\textit{id}), a string containing the question (\textit{question}), a paraphrased version of the question (\textit{paraphrased\_question}), a SPARQL query that answers the question (\textit{query}), the type of the query (\textit{query\_type}), the template of the query (\textit{template\_id}), a list of entities in the question (\textit{entities}), a list of relations in the question (\textit{relations}), a boolean indicating whether the question contains a temporal expression (\textit{temporal}) and a boolean indicating whether the question is held out from the training set (\textit{held\_out}). Sample data is shown in Figure 1.

\begin{figure*}[!ht]
    \centering
    \noindent{\includegraphics[width=\textwidth]{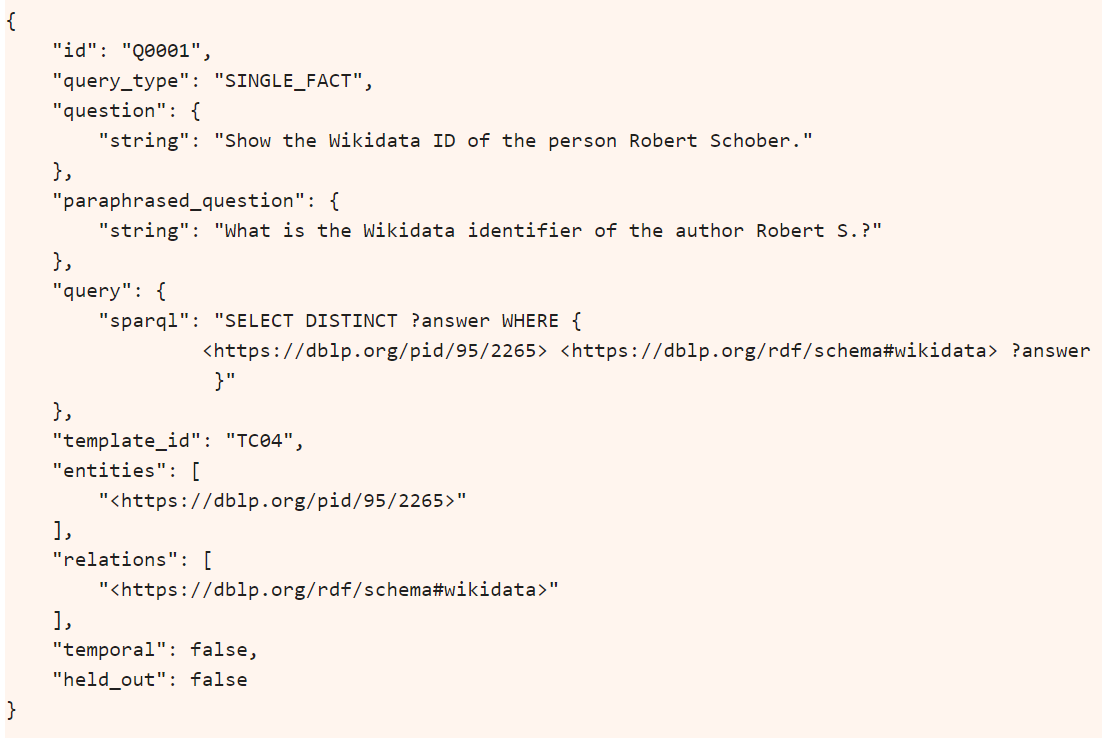}}
    \caption{Example of DBLP-QUAD data}
\end{figure*}

\subsubsection{Seed data}

Participants are provided with seed data to evaluate the performance of their systems on the QA and EL sub-tasks. This data is curated by the task organizers and consists of 500 random question-SPARQL pairs that comply with the same format as the DBLP-QUAD data. For the final evaluation, the organizers provided an additional set of 500 random questions containing only the question and its paraphrase. This dataset was used exclusively to score the performance of the participant systems in both sub-tasks.

\subsection{System description}

This section introduces our proposed system. It presents the system architecture and describes the training process in our experiments.

\subsubsection{PSYCHIC model}

Since the challenge presents itself as a QA task, we selected the DistilBERT-base uncased \footnote{https://huggingface.co/distilbert-base-uncased} model to tackle it in an extractive QA setting. Extractive QA confines a model to selecting the appropriate answer chunk from a context given as input with the question. It also leverages some control over the model answers as opposed to generative QA which makes models generate the answer.

Training a QA model involves giving the model an input composed of the question to be answered and a context that should contain the answer. In the scope of this shared task, we targeted two types of answers: the SPARQL query which should directly answer the given question and the list of entities for EL. We constructed the context to include both pieces of information and added information to guide the model regarding the nature of the question asked and the expected SPARQL answer: the query type and the template id. We also introduced symbolic information through a symbolic rule engine that inserts the special tokens [CLS] at the start of the context string and [SEP] between the different pieces of information in the context. These markers were added to ground the output of the model and enable it to discriminate between the different pieces of context information. The aim is to help the model learn the types of query structures and entities it will be asked to retrieve.

On the output side, we fine-tuned our PSYCHIC (\textbf{P}re-trained \textbf{SY}mbolic \textbf{CH}ecker \textbf{I}n \textbf{C}ontext) model to return both the SPARQL query and the list of entities. We constructed the output as a string containing both pieces of information separated by the [SEP] token. We also introduced another symbolic rule engine which is a programmatic function designed to split the output string based on the [SEP] symbol to return the query and entities chunks separately. The PSYCHIC model architecture is presented in Figure 2.

\begin{figure*}[!ht]
    \centering
    \noindent{\includegraphics[width=\textwidth]{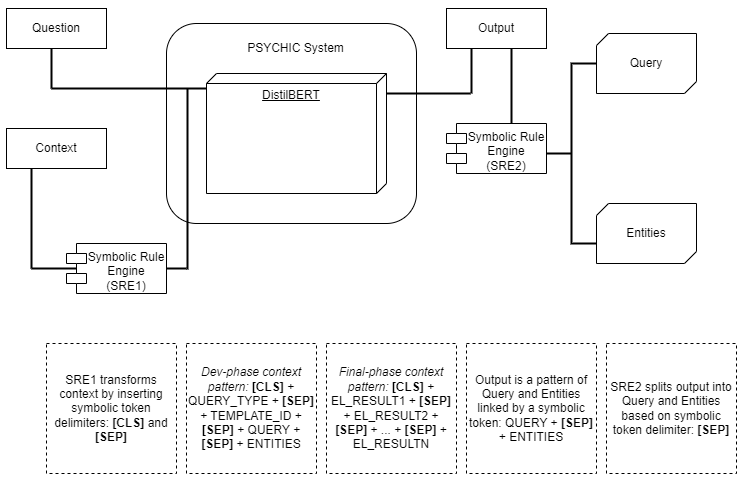}}
    \caption{PSYCHIC model architecture}
\end{figure*}

\subsubsection{Pipeline architecture}

To answer both parts of the shared task, we need to return an answer for the QA challenge and an entity list for the EL challenge. In that respect, the PSYCHIC model alone is not enough. The QA challenge defines an answer as the result of a SPARQL query, whereas PSYCHIC returns the queries themselves. This is a deliberate design choice since teaching a model to recognize patterned query structures is still an easier problem than teaching it to learn dynamic answers ranging from boolean values to lists of elements. Being able to correctly return the right query is essentially the hard part of the challenge, and the additional step to get the final answer consists in running the query returned by PSYCHIC using the SPARQL endpoint provided by the task organizers.

As such, we built the NS framework shown in Figure 3 to leverage the power of LLMs and symbolic reasoning. It takes as input the question-SPARQL pairs, processes them and prepares the question-context dataset needed for the PSYCHIC model. The model predicts the output in the form of a string containing the query, the separator [SEP] and the entity list. Two additional modules, the query and entity sanitizers, extract the relevant pieces of information from the output, namely the query and the entity list, by splitting the output string and validating the query and entity elements by matching them to their respective patterns. Through this sanitization process, these modules can perform error-correction to account for malformed strings returned by the model, e.g., transforming \textit{'select distinct? answer where \{? answer < https : / / dblp. org / rdf / schema \# authoredby > < https : / / dblp. org / pid / 00 / 2941 > \}'} to \textit{'select distinct ?answer where \{ ?answer <https://dblp.org/rdf/schema\#authoredBy> <https://dblp.org/pid/00/2941> \}'} for the query and \textit{['< https : / / dblp. org / pid / 00 / 2941 >']} to \textit{['<https://dblp.org/pid/00/2941>']} for the entity list.

The resulting output from the query sanitizer module is a correctly-formed SPARQL query that is run using the SPARQL endpoint to return the expected final answer. The output of the entity sanitizer is the correctly-formed entity list.

\begin{figure*}[!ht]
    \centering
    \noindent{\includegraphics[width=\textwidth]{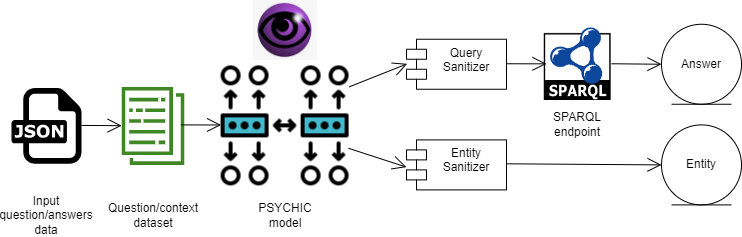}}
    \caption{Solution system architecture}
\end{figure*}

\subsubsection{Experimental setup}

We divided our experiment into two phases: training and inference or evaluation phase.

For the training phase, we treated the question and paraphrase as distinct questions to effectively double our overall training, validation and test sets. We trained the PSYCHIC model on the training and validation splits and evaluated on the test split. We set the following training hyperparameters for the model: learning rate = $2\mathrm{e}{-05}$, training batch size = 16, evaluation batch size = 16, seed = 42, optimizer = Adam with betas = (0.9, 0.999) and epsilon = $1\mathrm{e}{-08}$, and number of epochs = 3.

For the inference step, we distinguished two phases: the dev phase and the final phase. The dev phase corresponds to the phase when the first 500 random question-SPARQL pairs are made available by the organizers. The final phase represents the true system evaluation phase whereby the dataset used is the set of 500 random questions (and their paraphrases). The phases differ by the nature of the input given to the PSYCHIC model. In the dev phase, the context was constructed the same way as for DBLP-QUAD, i.e., following the [CLS] + QUERY\_TYPE + [SEP] + TEMPLATE\_ID + [SEP] + QUERY + [SEP] + ENTITIES pattern. In the final phase, since the only available information is the question and its paraphrase, we used an entity linker provided by the task organizers that leverages a t5-base language model with translating embeddings to return a list of predicted entities for each provided question. These results were concatenated using the symbolic rule engine to form a different context pattern, i.e., [CLS] + EL\_RESULT1 + [SEP] + EL\_RESULT2 + [SEP] + ... + [SEP] + EL\_RESULTN for N returned entity results.

We used the F1-QA and F1-EL metrics, representing the F1 scores on each of the QA and EL sub-tasks respectively, to evaluate our system. All experiments were performed on a Dell G15 Special Edition 5521 hardware with 14 CPU Cores, 32 GB RAM and NVIDIA GeForce RTX 3070 Ti GPU.

\section{Results \label{sec4}}
Table 1 displays the training results of the PSYCHIC model. Over 3 epochs, the model learns consistently with a perfect loss of 0 by the third epoch. Having integrated symbolic knowledge through the insertion of special tokens in the context proves to be a sound way to represent the different context chunks we want the model to learn. By the end of training, PSYCHIC can reliably return a complex output constituted of a query and an entity list for an input question.

The results of Table 2 seem to corroborate these findings. In the dev phase, the seed input structure is identical to that of DBLP-QUAD which means that PSYCHIC expects a similar context pattern. The model achieves a perfect score on both the QA and EL sub-tasks which suggests it is perfectly capable of discriminating query tokens from entity tokens. The results of the final phase are interesting in the sense that they seem to contradict this theory. The F1-QA score in particular suggests the model fails to predict any question query correctly. The reason for this bad performance lies in the fact that the input structure for the final phase set is radically different from the DBLP-QUAD and dev data as it contains no accompanying context for questions. Since PSYCHIC is an extractive QA model, it becomes very challenging for it to predict any kind of information without relevant context. We were not able to fill this context gap in the scope of this challenge.

For the EL sub-task, we were able to fill the gap by using the entity linker predictions as missing context for the input questions. Because these context patterns were unseen by PSYCHIC during the training and dev phases, the F1-EL score achieved by the model is impressive since it reveals it can correctly identify valid entity pattern structures. 

\begin{table}[ht]
\begin{center}
\begin{tabular}{|l|l|l|l|}
\hline \bf Epoch  & \bf Training Loss & \bf Step & \bf Validation Loss \\ 
\hline
1.0	& 0.001 & 1000 & 0.0001 \\ 
2.0	& 0.0005	& 2000	& 0.0000	\\
3.0	& 0.0002	& 3000	& 0.0000 \\
\hline
\end{tabular}
\end{center}
\caption{PSYCHIC training results}
\end{table}

\begin{table}[ht]
\begin{center}
\begin{tabular}{|l|l|l|l|l|l|l|l|l|}
\hline \bf Phase  & \bf F1-QA & \bf F1-EL \\ 
\hline
Dev	& 100.00	& 100.00	 \\ 
Final	&  00.18	& 71.00 \\
\hline
\end{tabular}
\end{center}
\caption{Final system evaluation performance}
\end{table}

\section{Conclusion \label{sec5}}
In this shared task, we propose a NS framework to tackle QA and EL sub-tasks over a KG. We explore the effects of including symbolic learning in the context of LLMs and evaluate the overall performance on the sub-tasks for a changing context. In the future, we plan to extend these symbolic mechanisms to generative models such as retrieval augmented generation (RAG) pipelines.

%%
%% Define the bibliography file to be used
\bibliography{sample-ceur}

\begin{thebibliography}{28}
\expandafter\ifx\csname natexlab\endcsname\relax\def\natexlab#1{#1}\fi
\providecommand{\url}[1]{\texttt{#1}}
\providecommand{\href}[2]{#2}
\providecommand{\path}[1]{#1}
\providecommand{\DOIprefix}{doi:}
\providecommand{\ArXivprefix}{arXiv:}
\providecommand{\URLprefix}{URL: }
\providecommand{\Pubmedprefix}{pmid:}
\providecommand{\doi}[1]{\href{http://dx.doi.org/#1}{\path{#1}}}
\providecommand{\Pubmed}[1]{\href{pmid:#1}{\path{#1}}}
\providecommand{\bibinfo}[2]{#2}
\ifx\xfnm\relax \def\xfnm[#1]{\unskip,\space#1}\fi
%Type = Article
\bibitem[{Peng et~al.(2023)Peng, Xia, Naseriparsa, and Osborne}]{peng2023knowledge}
\bibinfo{author}{C.~Peng}, \bibinfo{author}{F.~Xia}, \bibinfo{author}{M.~Naseriparsa}, \bibinfo{author}{F.~Osborne},
\newblock \bibinfo{title}{Knowledge graphs: Opportunities and challenges},
\newblock \bibinfo{journal}{Artificial Intelligence Review}  (\bibinfo{year}{2023}) \bibinfo{pages}{1--32}.
%Type = Article
\bibitem[{Banerjee et~al.(2023)Banerjee, Awale, Usbeck, and Biemann}]{banerjee2023dblp}
\bibinfo{author}{D.~Banerjee}, \bibinfo{author}{S.~Awale}, \bibinfo{author}{R.~Usbeck}, \bibinfo{author}{C.~Biemann},
\newblock \bibinfo{title}{Dblp-quad: A question answering dataset over the dblp scholarly knowledge graph},
\newblock \bibinfo{journal}{arXiv preprint arXiv:2303.13351}  (\bibinfo{year}{2023}).
%Type = Article
\bibitem[{Ishwari et~al.(2019)Ishwari, Aneeze, Sudheesan, Karunaratne, Nugaliyadde, and Mallawarrachchi}]{ishwari2019advances}
\bibinfo{author}{K.~Ishwari}, \bibinfo{author}{A.~Aneeze}, \bibinfo{author}{S.~Sudheesan}, \bibinfo{author}{H.~Karunaratne}, \bibinfo{author}{A.~Nugaliyadde}, \bibinfo{author}{Y.~Mallawarrachchi},
\newblock \bibinfo{title}{Advances in natural language question answering: A review},
\newblock \bibinfo{journal}{arXiv preprint arXiv:1904.05276}  (\bibinfo{year}{2019}).
%Type = Article
\bibitem[{Pan et~al.(2023)Pan, Razniewski, Kalo, Singhania, Chen, Dietze, Jabeen, Omeliyanenko, Zhang, Lissandrini et~al.}]{pan2023large}
\bibinfo{author}{J.~Z. Pan}, \bibinfo{author}{S.~Razniewski}, \bibinfo{author}{J.-C. Kalo}, \bibinfo{author}{S.~Singhania}, \bibinfo{author}{J.~Chen}, \bibinfo{author}{S.~Dietze}, \bibinfo{author}{H.~Jabeen}, \bibinfo{author}{J.~Omeliyanenko}, \bibinfo{author}{W.~Zhang}, \bibinfo{author}{M.~Lissandrini}, et~al.,
\newblock \bibinfo{title}{Large language models and knowledge graphs: Opportunities and challenges},
\newblock \bibinfo{journal}{arXiv preprint arXiv:2308.06374}  (\bibinfo{year}{2023}).
%Type = Article
\bibitem[{Zheng and Zhang(2019)}]{zheng2019question}
\bibinfo{author}{W.~Zheng}, \bibinfo{author}{M.~Zhang},
\newblock \bibinfo{title}{Question answering over knowledge graphs via structural query patterns},
\newblock \bibinfo{journal}{arXiv preprint arXiv:1910.09760}  (\bibinfo{year}{2019}).
%Type = Article
\bibitem[{Pramanik et~al.(2021)Pramanik, Alabi, Roy, and Weikum}]{pramanik2021uniqorn}
\bibinfo{author}{S.~Pramanik}, \bibinfo{author}{J.~Alabi}, \bibinfo{author}{R.~S. Roy}, \bibinfo{author}{G.~Weikum},
\newblock \bibinfo{title}{Uniqorn: unified question answering over rdf knowledge graphs and natural language text},
\newblock \bibinfo{journal}{arXiv preprint arXiv:2108.08614}  (\bibinfo{year}{2021}).
%Type = Article
\bibitem[{Sima et~al.(2022)Sima, Mendes~de Farias, Anisimova, Dessimoz, Robinson-Rechavi, Zbinden, and Stockinger}]{sima2022bio}
\bibinfo{author}{A.~C. Sima}, \bibinfo{author}{T.~Mendes~de Farias}, \bibinfo{author}{M.~Anisimova}, \bibinfo{author}{C.~Dessimoz}, \bibinfo{author}{M.~Robinson-Rechavi}, \bibinfo{author}{E.~Zbinden}, \bibinfo{author}{K.~Stockinger},
\newblock \bibinfo{title}{Bio-soda ux: enabling natural language question answering over knowledge graphs with user disambiguation},
\newblock \bibinfo{journal}{Distributed and Parallel Databases} \bibinfo{volume}{40} (\bibinfo{year}{2022}) \bibinfo{pages}{409--440}.
%Type = Article
\bibitem[{Zheng et~al.(2018)Zheng, Yu, Zou, and Cheng}]{zheng2018question}
\bibinfo{author}{W.~Zheng}, \bibinfo{author}{J.~X. Yu}, \bibinfo{author}{L.~Zou}, \bibinfo{author}{H.~Cheng},
\newblock \bibinfo{title}{Question answering over knowledge graphs: question understanding via template decomposition},
\newblock \bibinfo{journal}{Proceedings of the VLDB Endowment} \bibinfo{volume}{11} (\bibinfo{year}{2018}) \bibinfo{pages}{1373--1386}.
%Type = Inproceedings
\bibitem[{Nikas et~al.(2021)Nikas, Fafalios, and Tzitzikas}]{nikas2021open}
\bibinfo{author}{C.~Nikas}, \bibinfo{author}{P.~Fafalios}, \bibinfo{author}{Y.~Tzitzikas},
\newblock \bibinfo{title}{Open domain question answering over knowledge graphs using keyword search, answer type prediction, sparql and pre-trained neural models},
\newblock in: \bibinfo{booktitle}{The Semantic Web--ISWC 2021: 20th International Semantic Web Conference, ISWC 2021, Virtual Event, October 24--28, 2021, Proceedings 20}, \bibinfo{organization}{Springer}, \bibinfo{year}{2021}, pp. \bibinfo{pages}{235--251}.
%Type = Article
\bibitem[{Cm et~al.(2022)Cm, Prakash, and Singh}]{cm2022question}
\bibinfo{author}{S.~Cm}, \bibinfo{author}{J.~Prakash}, \bibinfo{author}{P.~K. Singh},
\newblock \bibinfo{title}{Question answering over knowledge graphs using bert based relation mapping},
\newblock \bibinfo{journal}{Expert Systems}  (\bibinfo{year}{2022}) \bibinfo{pages}{e13456}.
%Type = Article
\bibitem[{Diomedi and Hogan(????)}]{diomedi2107question}
\bibinfo{author}{D.~Diomedi}, \bibinfo{author}{A.~Hogan},
\newblock \bibinfo{title}{Question answering over knowledge graphs with neural machine translation and entity linking, 2021. doi: 10.48550},
\newblock \bibinfo{journal}{arXiv preprint arXiv.2107.02865}  (????).
%Type = Article
\bibitem[{Aghaei et~al.(2022)Aghaei, Raad, and Fensel}]{aghaei2022question}
\bibinfo{author}{S.~Aghaei}, \bibinfo{author}{E.~Raad}, \bibinfo{author}{A.~Fensel},
\newblock \bibinfo{title}{Question answering over knowledge graphs: A case study in tourism},
\newblock \bibinfo{journal}{IEEE Access} \bibinfo{volume}{10} (\bibinfo{year}{2022}) \bibinfo{pages}{69788--69801}.
%Type = Article
\bibitem[{Mavromatis and Karypis(2022)}]{mavromatis2022rearev}
\bibinfo{author}{C.~Mavromatis}, \bibinfo{author}{G.~Karypis},
\newblock \bibinfo{title}{Rearev: Adaptive reasoning for question answering over knowledge graphs},
\newblock \bibinfo{journal}{arXiv preprint arXiv:2210.13650}  (\bibinfo{year}{2022}).
%Type = Inproceedings
\bibitem[{Li et~al.(2021)Li, Wong, Fung, and Zhu}]{li2021improving}
\bibinfo{author}{S.~Li}, \bibinfo{author}{K.~W. Wong}, \bibinfo{author}{C.~C. Fung}, \bibinfo{author}{D.~Zhu},
\newblock \bibinfo{title}{Improving question answering over knowledge graphs using graph summarization},
\newblock in: \bibinfo{booktitle}{Neural Information Processing: 28th International Conference, ICONIP 2021, Sanur, Bali, Indonesia, December 8--12, 2021, Proceedings, Part IV 28}, \bibinfo{organization}{Springer}, \bibinfo{year}{2021}, pp. \bibinfo{pages}{489--500}.
%Type = Inproceedings
\bibitem[{Dutt et~al.(2022)Dutt, Bhattacharjee, Gangadharaiah, Roth, and Rose}]{dutt2022perkgqa}
\bibinfo{author}{R.~Dutt}, \bibinfo{author}{K.~Bhattacharjee}, \bibinfo{author}{R.~Gangadharaiah}, \bibinfo{author}{D.~Roth}, \bibinfo{author}{C.~Rose},
\newblock \bibinfo{title}{Perkgqa: Question answering over personalized knowledge graphs},
\newblock in: \bibinfo{booktitle}{Findings of the Association for Computational Linguistics: NAACL 2022}, \bibinfo{year}{2022}, pp. \bibinfo{pages}{253--268}.
%Type = Inproceedings
\bibitem[{Saxena et~al.(2020)Saxena, Tripathi, and Talukdar}]{saxena2020improving}
\bibinfo{author}{A.~Saxena}, \bibinfo{author}{A.~Tripathi}, \bibinfo{author}{P.~Talukdar},
\newblock \bibinfo{title}{Improving multi-hop question answering over knowledge graphs using knowledge base embeddings},
\newblock in: \bibinfo{booktitle}{Proceedings of the 58th annual meeting of the association for computational linguistics}, \bibinfo{year}{2020}, pp. \bibinfo{pages}{4498--4507}.
%Type = Article
\bibitem[{Zuo et~al.(2023)Zuo, Zhu, Wu, Wang, Qi, and Zhong}]{zuo2023improving}
\bibinfo{author}{Z.~Zuo}, \bibinfo{author}{Z.~Zhu}, \bibinfo{author}{W.~Wu}, \bibinfo{author}{W.~Wang}, \bibinfo{author}{J.~Qi}, \bibinfo{author}{L.~Zhong},
\newblock \bibinfo{title}{Improving question answering over knowledge graphs with a chunked learning network},
\newblock \bibinfo{journal}{Electronics} \bibinfo{volume}{12} (\bibinfo{year}{2023}) \bibinfo{pages}{3363}.
%Type = Article
\bibitem[{Rony et~al.(2022)Rony, Chaudhuri, Usbeck, and Lehmann}]{rony2022tree}
\bibinfo{author}{M.~R. A.~H. Rony}, \bibinfo{author}{D.~Chaudhuri}, \bibinfo{author}{R.~Usbeck}, \bibinfo{author}{J.~Lehmann},
\newblock \bibinfo{title}{Tree-kgqa: an unsupervised approach for question answering over knowledge graphs},
\newblock \bibinfo{journal}{IEEE Access} \bibinfo{volume}{10} (\bibinfo{year}{2022}) \bibinfo{pages}{50467--50478}.
%Type = Article
\bibitem[{Shi et~al.(2023)Shi, Yuan, Guo, Ma, Chen, and Zhang}]{shi2023knowledge}
\bibinfo{author}{J.~Shi}, \bibinfo{author}{Z.~Yuan}, \bibinfo{author}{W.~Guo}, \bibinfo{author}{C.~Ma}, \bibinfo{author}{J.~Chen}, \bibinfo{author}{M.~Zhang},
\newblock \bibinfo{title}{Knowledge-graph-enabled biomedical entity linking: a survey},
\newblock \bibinfo{journal}{World Wide Web}  (\bibinfo{year}{2023}) \bibinfo{pages}{1--30}.
%Type = Inproceedings
\bibitem[{Dubey et~al.(2018)Dubey, Banerjee, Chaudhuri, and Lehmann}]{dubey2018earl}
\bibinfo{author}{M.~Dubey}, \bibinfo{author}{D.~Banerjee}, \bibinfo{author}{D.~Chaudhuri}, \bibinfo{author}{J.~Lehmann},
\newblock \bibinfo{title}{Earl: joint entity and relation linking for question answering over knowledge graphs},
\newblock in: \bibinfo{booktitle}{The Semantic Web--ISWC 2018: 17th International Semantic Web Conference, Monterey, CA, USA, October 8--12, 2018, Proceedings, Part I 17}, \bibinfo{organization}{Springer}, \bibinfo{year}{2018}, pp. \bibinfo{pages}{108--126}.
%Type = Inproceedings
\bibitem[{Steinmetz(2023)}]{steinmetz2023entity}
\bibinfo{author}{N.~Steinmetz},
\newblock \bibinfo{title}{Entity linking for kgqa using amr graphs},
\newblock in: \bibinfo{booktitle}{European Semantic Web Conference}, \bibinfo{organization}{Springer}, \bibinfo{year}{2023}, pp. \bibinfo{pages}{122--138}.
%Type = Inproceedings
\bibitem[{Radhakrishnan et~al.(2018)Radhakrishnan, Talukdar, and Varma}]{radhakrishnan2018elden}
\bibinfo{author}{P.~Radhakrishnan}, \bibinfo{author}{P.~Talukdar}, \bibinfo{author}{V.~Varma},
\newblock \bibinfo{title}{Elden: Improved entity linking using densified knowledge graphs},
\newblock in: \bibinfo{booktitle}{Proceedings of the 2018 Conference of the North American Chapter of the Association for Computational Linguistics: Human Language Technologies, Volume 1 (Long Papers)}, \bibinfo{year}{2018}, pp. \bibinfo{pages}{1844--1853}.
%Type = Misc
\bibitem[{Thawani et~al.(????)Thawani, Hu, Hu, Zafar, Divvala, Singh, Qasemi, Szekely, and Pujara}]{thawanientity}
\bibinfo{author}{A.~Thawani}, \bibinfo{author}{M.~Hu}, \bibinfo{author}{E.~Hu}, \bibinfo{author}{H.~Zafar}, \bibinfo{author}{N.~Divvala}, \bibinfo{author}{A.~Singh}, \bibinfo{author}{E.~Qasemi}, \bibinfo{author}{P.~Szekely}, \bibinfo{author}{J.~Pujara}, \bibinfo{title}{Entity linking to knowledge graphs to infer column types and properties, 2019, semantic web challenge on tabular data to knowledge graph matching, iswc}, ????
%Type = Article
\bibitem[{Li et~al.(2022)Li, Li, Li, Li, Liu, Liu, and Dong}]{li2022improving}
\bibinfo{author}{Q.~Li}, \bibinfo{author}{F.~Li}, \bibinfo{author}{S.~Li}, \bibinfo{author}{X.~Li}, \bibinfo{author}{K.~Liu}, \bibinfo{author}{Q.~Liu}, \bibinfo{author}{P.~Dong},
\newblock \bibinfo{title}{Improving entity linking by introducing knowledge graph structure information},
\newblock \bibinfo{journal}{Applied Sciences} \bibinfo{volume}{12} (\bibinfo{year}{2022}) \bibinfo{pages}{2702}.
%Type = Article
\bibitem[{Huang et~al.(2020)Huang, Wang, Wang, Liu, and Liu}]{huang2020entity}
\bibinfo{author}{B.~Huang}, \bibinfo{author}{H.~Wang}, \bibinfo{author}{T.~Wang}, \bibinfo{author}{Y.~Liu}, \bibinfo{author}{Y.~Liu},
\newblock \bibinfo{title}{Entity linking for short text using structured knowledge graph via multi-grained text matching}  (\bibinfo{year}{2020}).
%Type = Inproceedings
\bibitem[{Banerjee et~al.(2020)Banerjee, Chaudhuri, Dubey, and Lehmann}]{banerjee2020pnel}
\bibinfo{author}{D.~Banerjee}, \bibinfo{author}{D.~Chaudhuri}, \bibinfo{author}{M.~Dubey}, \bibinfo{author}{J.~Lehmann},
\newblock \bibinfo{title}{Pnel: Pointer network based end-to-end entity linking over knowledge graphs},
\newblock in: \bibinfo{booktitle}{The Semantic Web--ISWC 2020: 19th International Semantic Web Conference, Athens, Greece, November 2--6, 2020, Proceedings, Part I 19}, \bibinfo{organization}{Springer}, \bibinfo{year}{2020}, pp. \bibinfo{pages}{21--38}.
%Type = Inproceedings
\bibitem[{Ding et~al.(2021)Ding, Chaudhri, Chittar, and Konakanchi}]{ding2021jel}
\bibinfo{author}{W.~Ding}, \bibinfo{author}{V.~K. Chaudhri}, \bibinfo{author}{N.~Chittar}, \bibinfo{author}{K.~Konakanchi},
\newblock \bibinfo{title}{Jel: applying end-to-end neural entity linking in jpmorgan chase},
\newblock in: \bibinfo{booktitle}{Proceedings of the AAAI Conference on Artificial Intelligence}, volume~\bibinfo{volume}{35}, \bibinfo{year}{2021}, pp. \bibinfo{pages}{15301--15308}.
%Type = Inproceedings
\bibitem[{Diomedi and Hogan(2022)}]{diomedi2022entity}
\bibinfo{author}{D.~Diomedi}, \bibinfo{author}{A.~Hogan},
\newblock \bibinfo{title}{Entity linking and filling for question answering over knowledge graphs},
\newblock in: \bibinfo{booktitle}{Natural Language Interfaces for the Web of Data (NLIWOD) Workshop}, \bibinfo{year}{2022}.

\end{thebibliography}


\end{document}